\documentclass[conference]{IEEEtran}
\IEEEoverridecommandlockouts
\usepackage{cite}
\usepackage{amsmath,amssymb,amsfonts}
\usepackage{algorithmic}
\usepackage{graphicx}
\usepackage{mathtools}
\usepackage{textcomp}
\usepackage{xcolor}
\usepackage{url} 
\def\BibTeX{{\rm B\kern-.05em{\sc i\kern-.025em b}\kern-.08em
    T\kern-.1667em\lower.7ex\hbox{E}\kern-.125emX}}
    
\begin{document}

\title{Trip Planning for Autonomous Vehicles with Wireless Data Transfer Needs Using Reinforcement Learning\\}

\author{\IEEEauthorblockN{Yousef AlSaqabi*}

\IEEEauthorblockA{\textit{Viterbi School of Engineering} \\
\textit{University of Southern California}\\
Los Angeles, CA, United States \\
alsaqabi@usc.edu}

{\footnotesize \textsuperscript{*}This author is also affiliated with Kuwait University as a PhD scholar.}

\and
\IEEEauthorblockN{Bhaskar Krishnamachari}
\IEEEauthorblockA{\textit{Viterbi School of Engineering} \\
\textit{University of Southern California}\\
Los Angeles, CA, United States \\
bkrishna@usc.edu}
}
\maketitle
\begin{abstract}

With recent advancements in the field of communications and the Internet of Things, vehicles are becoming more aware of their environment and are evolving towards full autonomy.  
Vehicular communication opens up the possibility for vehicle-to-infrastructure interaction, where vehicles could share information with components such as cameras, traffic lights, and signage that support a country’s road system. 
As a result, vehicles are becoming more than just a means of transportation; they are collecting, processing, and transmitting massive amounts of data used to make driving safer and more convenient. 
With 5G cellular networks and beyond, there is going to be more data bandwidth available on our roads, but it may be heterogeneous because of limitations like line of sight, infrastructure, and heterogeneous traffic on the road.
This paper addresses the problem of route planning for autonomous vehicles in urban areas accounting for both driving time and data transfer needs. We propose a novel reinforcement learning solution that prioritizes high bandwidth roads to meet a vehicle's data transfer requirement, while also minimizing driving time. We compare this approach to traffic-unaware and bandwidth-unaware baselines to show how much better it performs under heterogeneous traffic. This solution could be used as a starting point to understand what good policies look like, which could potentially yield faster, more efficient heuristics in the future.

\end{abstract}

\begin{IEEEkeywords}
Vehicular networks, route planning, reinforcement learning
\end{IEEEkeywords}

\section{Introduction}

\subsection{Background and Motivation}

With recent advancements in the field of communications and the Internet of Things (IoT), vehicles are evolving towards full autonomy and are becoming more aware of their environment. IoT has become a leading technology that is constantly evolving to include new and improved devices and services like vehicular communication~\cite{IoV}, which is regarded as an important component of the development of the intelligent transportation system (ITS) and smart cities.

Vehicular communication opens up the possibility for vehicle-to-infrastructure (V2I) interaction, where vehicles could share information with components such as cameras, traffic lights, and signage that support a country’s road system~\cite{Ibanez}. For example, vehicles could interact with roadside units with dedicated computing resources to offload computing processes from vehicles, or to perform real-time calibration/tuning functions while they are en route~\cite{Taha}. As a result, vehicles are becoming more than just a means of transportation; they are generating, collecting, storing, processing, and transmitting massive amounts of data used to make driving safer and more convenient~\cite{Hao}. 

With 5G cellular networks and beyond, there is going to be more data bandwidth available on our roads, but it may be heterogeneous because of limitations like line of sight, infrastructure, and heterogeneous traffic on the road. A lot of prior work has focused on intelligent route planning for autonomous vehicles~\cite{Bast}, but relatively less on route planning with heterogeneous bandwidth and communication requirements taken into account. This paper addresses the problem of route planning for autonomous vehicles in urban areas accounting for both drive time and the data upload/download needs.

\subsection{Prior and Related Work}

Route planning defines the waypoints of the journey based on the map of the road network according to some evaluation standards such as shortest or fastest path~\cite{V2V}. Deterministic algorithms are traditionally used for route planning purposes~\cite{Aradi}. A deterministic planning process is one in which the available information is static and assumed to be complete. A deterministic algorithm can select an entire route from a map that extends from a starting point to an end point. It can specify an entire route that gives the shortest path beforehand, and it doesn't need more information to be obtained as the plan is executed. The A* algorithm is one of the most efficient algorithms for calculating a safe route with the shortest distance cost~\cite{A*}. Multi-agent path planning (MAPP) is increasingly being used to address resource allocation problems in dynamic, distributed environments that involve autonomous agents~\cite{MAPP}. Many other intelligent algorithms such as fuzzy logic~\cite{Fuzzy}, generic algorithm (GA)~\cite{GA}, and neural networks~\cite{NN} have been used for path planning of mobile robots. 

Machine learning (ML), a major branch of artificial intelligence, builds intelligent systems to operate in complex environments, and has found many successful applications in many fields including computer vision, medical diagnosis, search engines, and robotics~\cite{ML}. Machine learning techniques can generally be divided into supervised learning, unsupervised learning, and reinforcement learning. Supervised learning is a ML method where models are trained using labeled datasets that ``supervise" it to help predict outcomes or unforeseen data. In unsupervised learning, there is no need to label the data; the model is allowed to work on its own to discover information. Reinforcement learning (RL), is a family of algorithms which allows agents to learn how to act in different situations; it describes how to establish a policy, which is a map from observations (states) to actions, that maximize a numerical reward signal~\cite{RL}. In RL, the agent continuously interacts with a dynamic environment and tries to develop a good strategy based on the reward/cost of the environment's feedback. Deep Reinforcement learning (DRL)~\cite{DRL} is an approach that extends RL by using a deep neural network to estimate the states instead of having to map every solution, creating a more manageable solution space in the decision process. 

RL/DRL have been recently used in vehicular networks as an evolving method to solve different problems and challenges effectively. It is used for applications like autonomous highway driving where the vehicle learns to make decisions by interacting with the surrounding traffic~\cite{Nageshrao}. It is also used for vehicle motion planning purposes where the vehicle learns different tasks like car following, lane merging, lane keeping, and overtaking~\cite{motionplanning}. Other than motion planning, it is also used for adaptive computation offloading in heterogeneous vehicular networks~\cite{Ke9091251}

When it comes to route planning, RL/DRL could be preferable to deterministic algorithms because of uncertainties in traffic conditions (from hour to hour or day to day) that would also affect the bandwidth conditions on the roads~\cite{Planning}. DRL has been successfully used for route planning and navigation purposes under realistic traffic simulations~\cite{KOH2020106694}. It can also be used for specific applications of route planning that focus on minimizing energy consumption in plug-in hybrid electric vehicles~\cite{powermgmt}. 

Other than applications for urban cities, RL has been used for robot~\cite{robot} and multi-robot~\cite{multirobot} route planning optimization. DRL is also used for route planning applications for unmanned aerial vehicle (UAV)-mounted mobile edge computing networks can also benefit from using DRL~\cite{uav}.

\subsection{Paper Contributions}

A lot of prior work on autonomous vehicles has focused on using RL for intelligent motion and route planning. RL has been successfully used for the route planning of robots~\cite{robot}, multi-robots~\cite{multirobot}, and UAV's~\cite{uav} with goals of optimizing drive time, avoiding traffic~\cite{Nageshrao}, and minimizing energy consumption~\cite{powermgmt}. It has also shown to be useful for adaptive computational offloading in heterogeneous vehicular networks~\cite{Ke9091251}. However, relatively less work has focused on addressing our problem of route planning with heterogeneous bandwidth and communication requirements taken into account. In this paper, RL is proposed as a novel solution to this problem. It could also be used as a starting point to understand what good policies look like, which could potentially yield faster, more efficient heuristics in the future.

The contributions of this paper are as follows: 

\begin{itemize}
    \item We formulate the problem of route determination considering both traffic and bandwidth heterogeneity as a reinforcement learning problem; to our knowledge, this is the first such formulation in the literature.
    \item We implement an environment using OpenAI Gym for the problem using real map data on vehicle mobility in a city to evaluate the reinforcement learning approach.
    \item Using different deep reinforcement learning algorithms, we evaluate the RL approach with respect to various parameters pertaining to the reward function, the map, communication bandwidth, and  data requirements.
    \item We also evaluate the RL approach in comparison with alternative baselines such as traffic-unaware and bandwidth-unaware alternatives.
    \item We show that the RL approach is able to learn good solutions efficiently compared to baselines. 
\end{itemize}

\subsection{Paper Outline}

The remainder of this paper is organized as follows: Section \ref{formulation} gives a detailed description of the problem formulation and describes our RL definition. Section \ref{simulation} describes our simulation set-up, explaining how we used our dataset, the different map and RL parameters chosen, and details how we implemented OpenAI Gym to solve the problem.. In Section \ref{results}, experimental results are presented to demonstrate the episode completion time of the RL agents and compare them under different RL parameters. Section \ref{conclusion} presents the conclusive remarks and suggests future work. 

\section{Problem Formulation} \label{formulation}

\subsection{Problem Description} \label{2a}

\begin{figure}[!h]
    \centering
    \graphicspath{ {./Figures/} }
    \includegraphics [scale = 0.17] {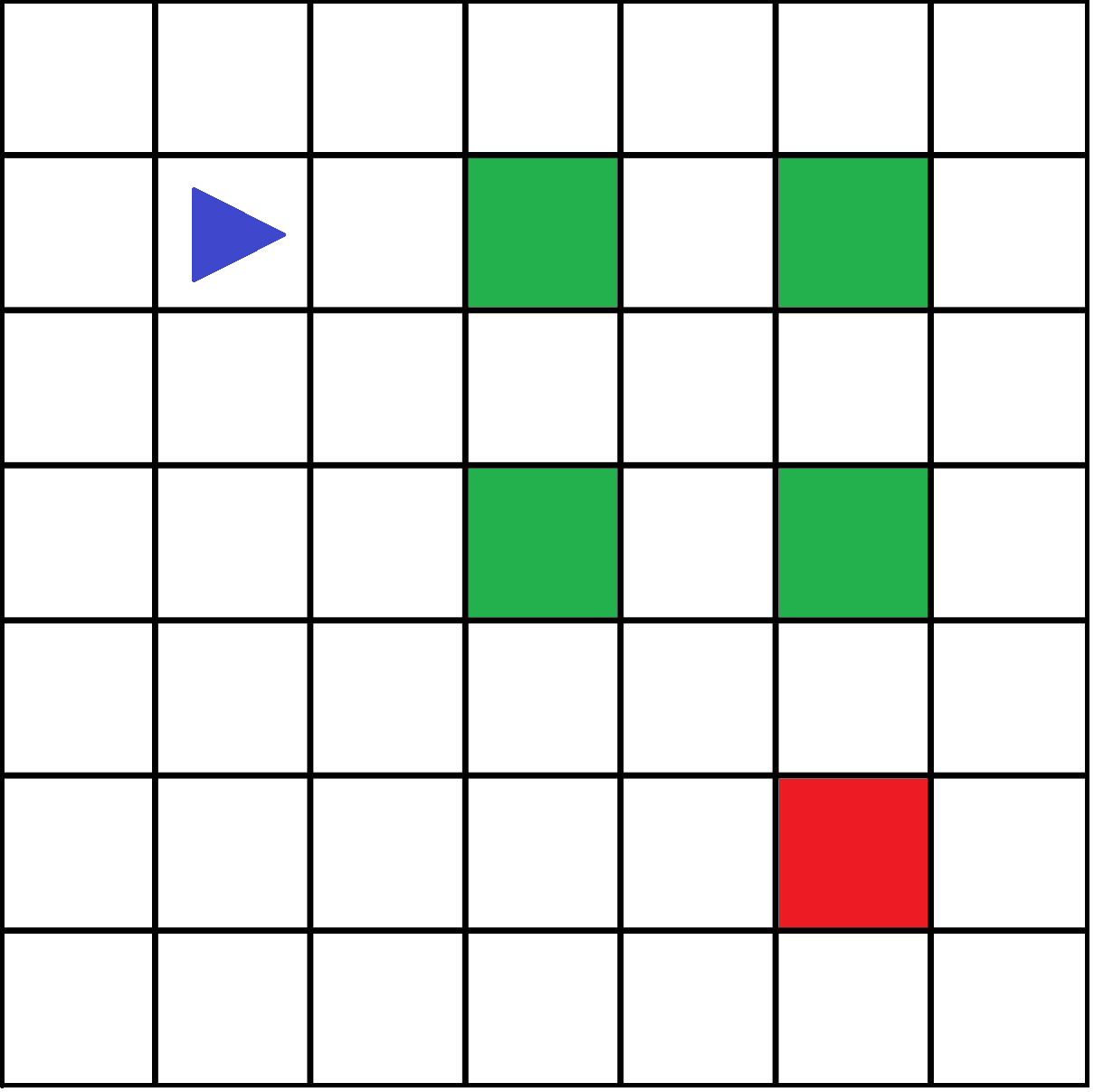}
    \caption{Grid Environment Example}
    \label{fig:env}
\end{figure}

The environment for this reinforcement learning problem is constructed as a 2D grid world environment that simulates an agent (vehicle) navigating through a city. NovelGridworlds~\cite{gridworlds} is the OpenAI Gym framework used to create this environment. Each cell in the grid represents a road stretch that can be traversed in one of four directions: north, south, east, and west. Cells in the grid will have binary bandwidth values; high bandwidth cells represent a road stretch that has network coverage from a nearby base station, and low bandwidth cells represent a road stretch with little to no coverage. One of the cells in the environment will also represent the final destination. An example of our environment can be seen in Fig.~\ref{fig:env}, where the green blocks are the high bandwidth cells, the red block is the final destination, and the blue triangle is the agent.

Each cell in the grid will have a value for traffic density, for example taken from the Uber movement dataset (more on this in Section \ref{simulation} below). To model the relationship between traffic density and agent speed, we will categorize the traffic density values into four different groups that would resemble realistic traffic conditions on the road: high density, medium-high density, medium-low density, and low density. The high density category represents traffic exceeding the road capacity (slow, bumper-to-bumper), the medium density categories represent traffic flowing at capacity, and the low density represents traffic flowing at less than capacity (when there are little to no cars on the road). Each category will have its own constant called the delay index that would represent how much time the agent spends in each cell. The relationship between traffic density and agent speed is step-wise decreasing; a higher traffic category would mean that it would take longer for the agent to drive through the cell.

The agent in the environment will have a certain upload/download requirement to meet before reaching the end destination. Every time the agent visits a high bandwidth cell, it can work towards that requirement; the data to be uploaded/downloaded when visiting a high bandwidth cell can be modeled by [\(\frac{1}{density} * delay  index\)], which is the bandwidth per vehicle multiplied by the time spent in the cell. The agent cannot complete the episode by reaching the end destination without visiting enough high bandwidth cells and meeting the requirement first.

\subsection{Reinforcement Learning Formulation}

For our problem, information about the state of the environment is not lost, but retained and sensed with each action. Since our problem exhibits this Markov Property, we can formulate it as a Markov Decision Process (MDP). 

\vspace{2mm} 

\begin{enumerate}
    \item \textbf{Policy ($\pi$)}: The ideal path the agent can take that maximizes the reward, so that the agent can reach the destination in the shortest time while also transferring all its data
    \item \textbf{Goal}: To find the optimal policy $\pi$*
    \item \textbf{State} (S): The state of the environment is described by the following parameters:
    \begin{itemize}
        \item Bandwidth of the cells in the grid
        \item Current position of the agent
        \item Amount of data left to transfer 
        \end{itemize}
    \vspace{1mm} 
    
    Bandwidth conditions in the grid can be represented by binary vectors, where each element in the vector represents a cell in the grid, and each cell can be either high bandwidth (1) or low bandwidth (0). Therefore, there could be a maximum of $2^N$ possible binary vectors, where N is the number of cells in the grid. The agent can be in any one of the N cells at a time, and the amount of data left to transfer can be denoted by \emph{(dt)}. Therefore, the total possible number of states, $S_{tot}$ is given by equation~\ref{states}
    
    \begin{equation} \label{states}
    S_{tot} = 2^N * N * dt
    \end{equation}
    
    \item \textbf{Action (A(s))}: From any given state S, the agent can choose to move one cell forward, right, or left
    
    \item \textbf{Reward ($R_{SS'}^a$)}: A numerical value given to the agent when transitioning from state S to state S' upon action a.
    \vspace{2mm} 
    
    The agent will be trained on two different reward functions, the step reward function and the cumulative reward function, to see which one performs better. In the step reward function, the agent is given a reward of +1 for every high bandwidth cell visited until meeting the bandwidth requirement, and +10 for reaching the final destination after meeting the bandwidth requirement. In the cumulative reward function, the agent is given a reward of +3 only after meeting the bandwidth requirement, and +10 for reaching the final destination after meeting that requirement. In both reward functions, the agent is given a negative reward of -[\(\frac{density}{mean density}\)]. The negative reward, or punishment, is scaled off the traffic density in each cell: the higher the traffic density in a cell, the bigger value for the punishment. The denominator in this punishment is the mean traffic density of all cells, and is used as a reference level to scale the punishment down to a suitable range. These reward functions train the agent so that it prioritizes high bandwidth cells to meet the data transfer requirements before reaching the destination while simultaneously taking the route with the least traffic.
    
    \vspace{1mm} 
    
    \item \textbf{Transition Probability ($P_{SS'}^a$)}: The probability that an action a in state s at time t will lead to state s' at time t+1 is given by equation~\ref{prob}
    
    \begin{equation} \label{prob}
    P_{SS'}^a = Pr(s_{t+1} = s' | s_t = s, a_t = a)
    \end{equation}
    
    \item \textbf{Discount ($\gamma$)}: $\gamma$ $\in$ [0,1] is the discount factor which represents the difference in importance between future rewards and present rewards. Since we are dealing with episodic tasks with a clear terminal state, we set this value to be 1.
\end{enumerate}

\section{Simulation Set-Up} \label{simulation}

\subsection {Traffic Dataset}

\begin{figure*}[ht!]
            \includegraphics[width=.3\textwidth]{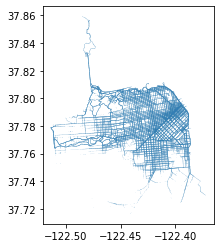}\hfill
            \includegraphics[width=.3\textwidth]{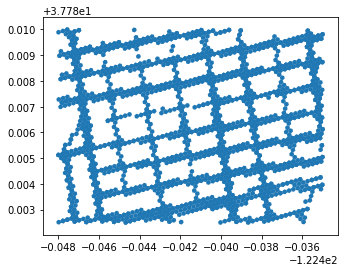}\hfill
            \includegraphics[width=.35\textwidth]{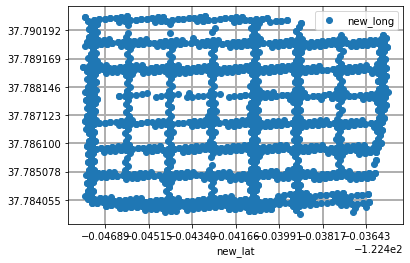}
            \caption{Dataset Cleaning Process}
            \label{fig:cleaning}
        \end{figure*}

Real world traffic data is used from Uber Movement’s mobility heatmap~\cite{Uber}. The heatmap’s downloadable CSV is generated by aggregating all GPS pings by day type (either weekend or weekday). It counts the number of traversals in a given location, so if there is a series of 2 or more pings in a row from the same hex, only the first ping in the series would be counted. These data points can also be filtered by any specified time interval throughout the week. 

To get our simulation to be as realistic as possible, we use the dataset for downtown San Francisco. The city’s streets all have the same speed limits and are already arranged in a grid that closely resembles our OpenAI gym grid environment. 

The Uber movement dataset gives us the traffic heatmap for the whole city which we don’t really need. Fig.~\ref{fig:cleaning}  highlights the data cleaning process. We first zoom in on a particular area of interest, which in this case is a part of the city that is already in a grid. We then rotate the plot so we can have the road stretches either vertical or horizontal. Lastly, we divide up the plot into a grid like we have in our environment. Summing up the traversals in each cell gives us the traffic density value for that cell.

The traffic values (traversals) from this dataset go up to 4000 traversals per cell, and we use these values to set appropriate values for the delay index of each traffic density category. As described in Section \ref{2a}, The relationship between traffic density and agent speed is step-wise decreasing; agents in a higher traffic category would take longer to drive through the cell. We assume that an agent takes one unit of time, or a delay index of 1, to drive through a cell with no other cars on the road. The delay index values for the different traffic groups are chosen based off how much more time would be needed to drive through the cell with that amount of traffic compared to no traffic at all. For example, a traffic group with a delay index of 3 suggests that it takes the agent triple the amount of time to drive through compared to if there were no other cars on the road.

\begin{figure}[!h]
    \centering
    \graphicspath{ {./Figures/} }
    \includegraphics [scale = 0.11] {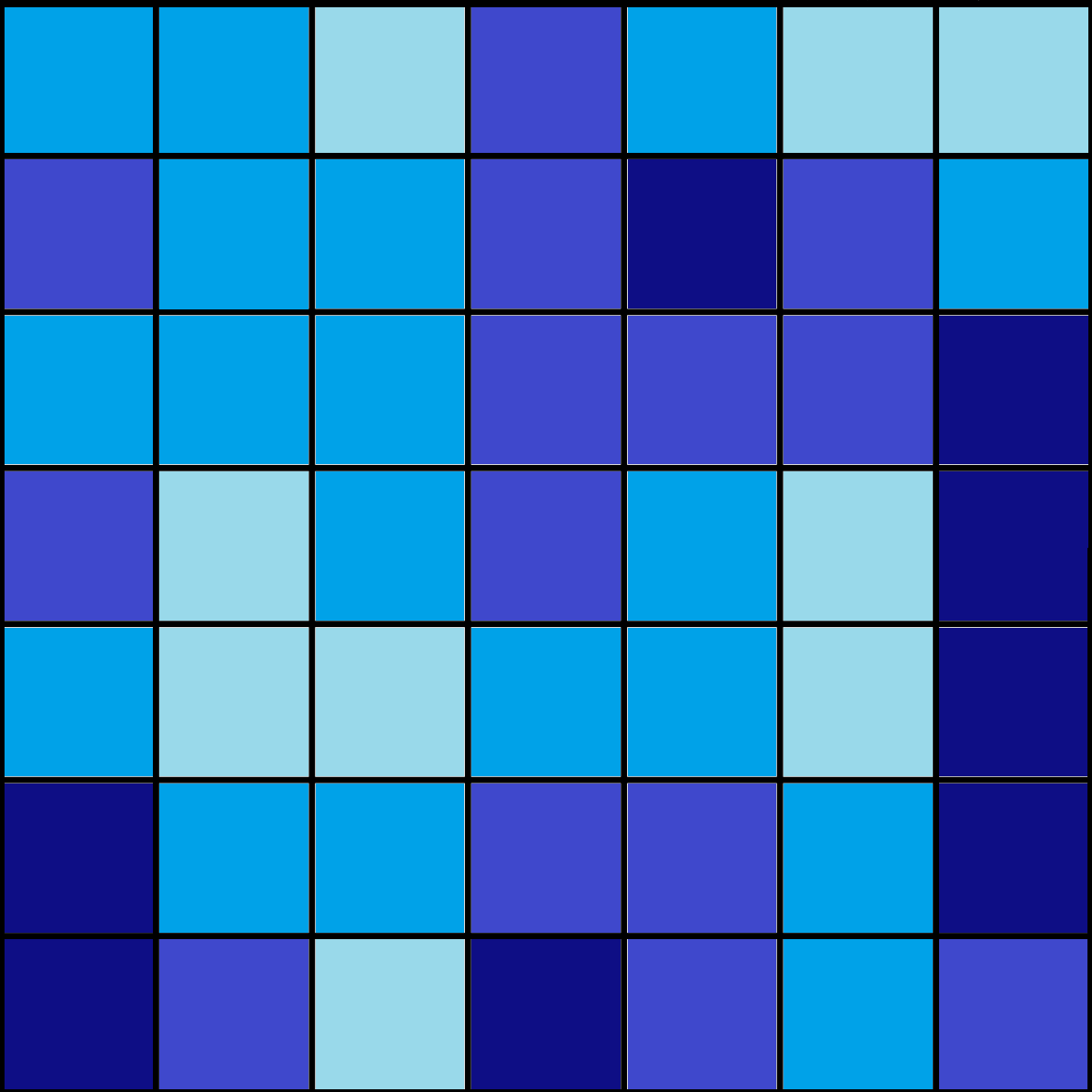}
    \caption{Traffic Density Heatmap}
    \label{fig:heatmap}
\end{figure}

The high traffic density group will have a delay index of 3 and will include cells with traffic values from 2900 and beyond. The medium-high traffic density group will have a delay index value of 2.5 and will include cells with traffic values between 2200 and 2900. The medium-low traffic density group will have a delay index value of 2 and will include cells with traffic values between 1100 and 2200. The low traffic density group will have a delay index value of 1.5 and will include cells with traffic values up to 1100. Fig.~\ref{fig:heatmap} shows the traffic density heatmap for our grid; four shades of blue correspond to the four different traffic density groups, with the darkest shade representing the most dense traffic group. 

\subsection {Simulation Parameters}

\begin{figure}[!h]
    \centering
    \graphicspath{ {./Figures/} }
    \includegraphics [scale = 0.35] {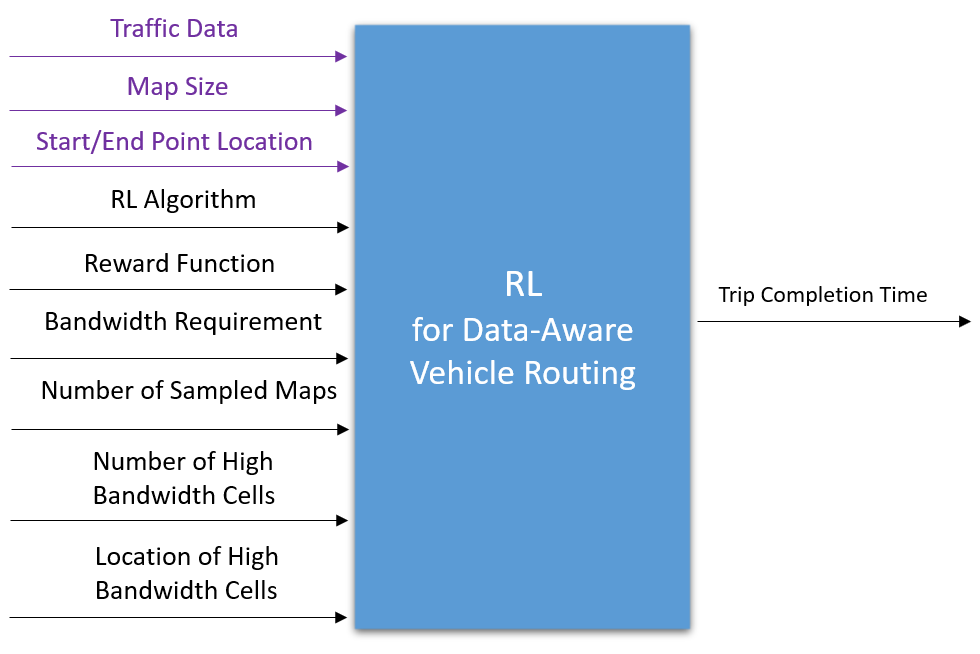}
    \caption{Simulation Parameters}
    \label{fig:block}
\end{figure}

Fig.~\ref{fig:block} shows the different parameters that contribute to the agent's trip completion time. The top three colored parameters represent the parameters that stay constant throughout the simulation; for every experiment, the same set of traffic data is used, and the map size and start/end locations are fixed. To explore the impact of a specific parameter, we vary that parameter while fixing the values for all the other parameters.

For this simulation, we compare the two reward functions (step and cumulative) on different map parameters in order to find out which function performs better. We fix our start and end points and set the map size to be a 7x7 grid. We then train the two reward functions while varying parameters like number of high bandwidth cells, their locations, the agent's bandwidth requirement, the number of maps to be sampled, and also the RL algorithm used.

\begin{figure}[!h]
    \centering
    \graphicspath{ {./Figures/} }
    \includegraphics [scale = 0.24] {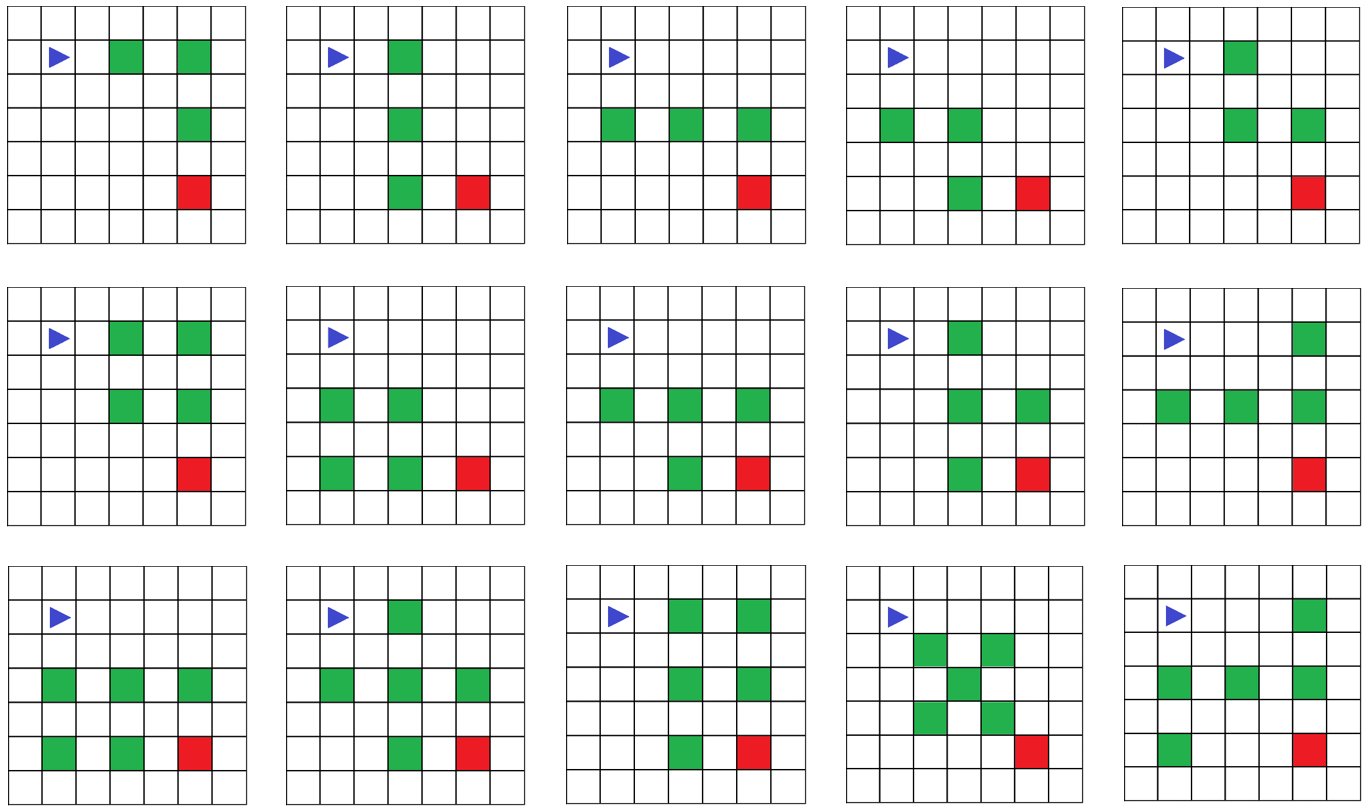}
    \caption{High Bandwidth Allocations in Different Maps}
    \label{fig:maps}
\end{figure}

Fig.~\ref{fig:maps} shows the different bandwidth allocations for a 7x7 grid with a fixed start and end point where the number of high bandwidth cells and the number of maps to be sampled are varied. The blue triangle represents the agent's starting point, the green blocks represent the high bandwidth cells, and the red block represents the end destination. The first row shows five unique maps with three high bandwidth cells in each. The second and third row also show five unique maps, but with four and five high bandwidth cells respectively. These are the maps that are used throughout our experiments in Section~\ref{results}.

\subsection{OpenAI Gym Implementation} \label{gymimp}

OpenAI Gym is a toolkit for developing and comparing reinforcement learning algorithms~\cite{openaigym}. The gym library comes with a diverse suite of environments ranging from very simple games to complex physics-based engines. It also allows users to create custom environments that can be used to work out RL algorithms. OpenAI's other package, Baselines, is a set of high-quality implementations of RL algorithms~\cite{baselines} that provides a simple interface to train RL agents and benchmark different RL algorithms. OpenAI Gym is used to develop the 2D grid world environment mentioned above.

In the next section, we compare three different RL algorithms from Stable Baselines~\cite{sb}, which is a fork from the OpenAI Baselines of RL algorithms: PPO2, DQN, and A2C. The Proximal Policy Optimization (PPO2) algorithm is a new family of policy gradient methods for RL which alternates between sampling data through interaction with the environment, and optimizing ``surrogate" objective function using stochastic gradient descent~\cite{ppo}. The Deep Q Network (DQN)~\cite{dqnAtari} algorithm is a variation of the classic Q-Learning algorithm with a deep convolutional neural net architecture for Q-function approximation and uses older network parameters to estimate the Q-values of the next state~\cite{dqn}. The Advantage Actor Critic (A2C) algorithm is a synchronous, deterministic implementation that waits for each actor to finish its segment of experience before updating, averaging over all of the actors~\cite{a2c}. In Section~\ref{results}, we will be training the model using these three algorithms with their default hyperparameters.

\section{Experiments and Results} \label{results}

\subsection {Reinforcement Learning Algorithm Comparison}

\begin{figure}[!h]
    \centering
    \graphicspath{ {./Figures/} }
    \includegraphics [width = 85mm]{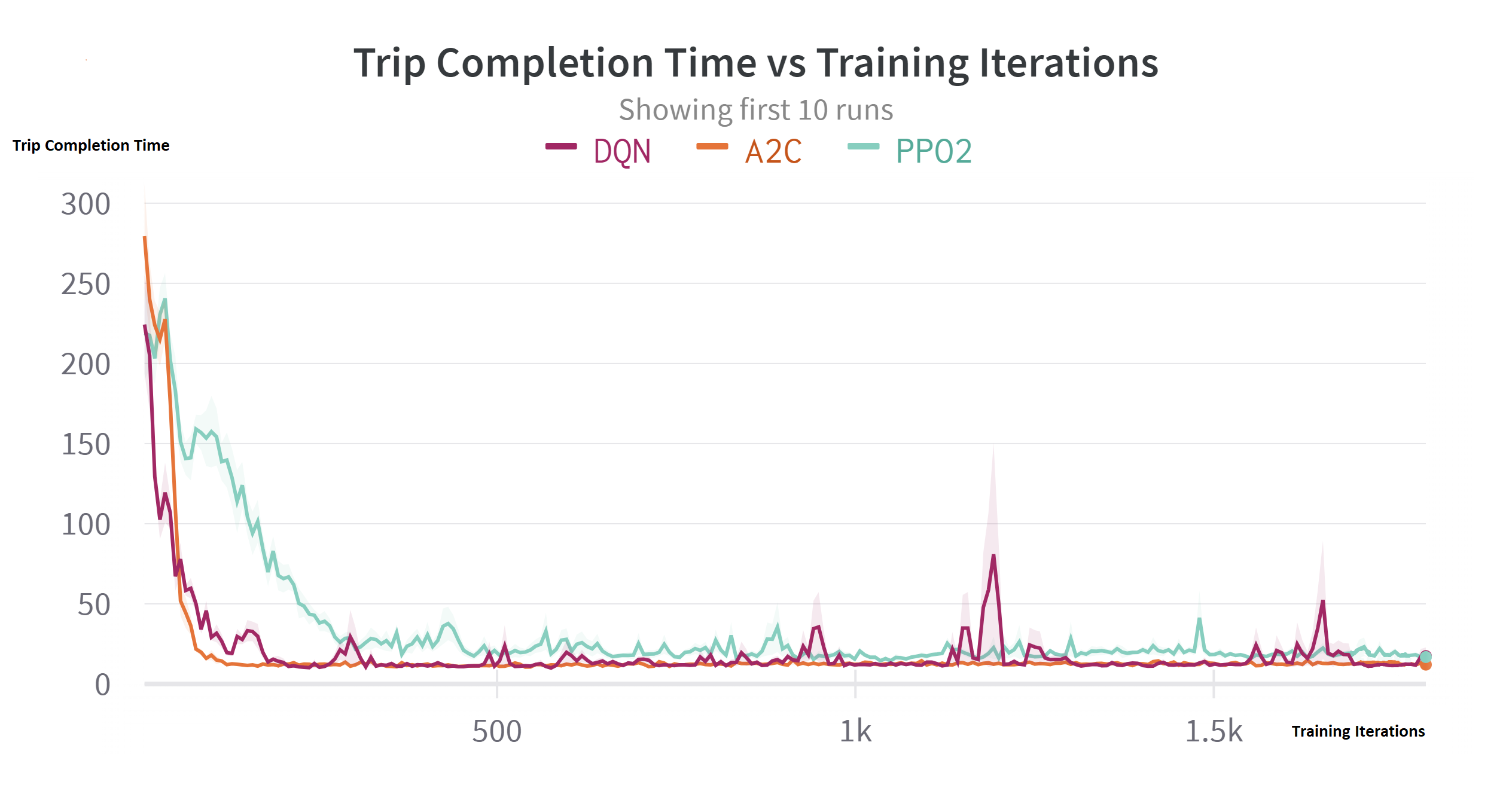}
    \caption{Comparing Different RL Algorithms}
    \label{fig:algocomp}
\end{figure}

Our first experiment is to compare the performance of three different RL algorithms mentioned previously in Section~\ref{gymimp}: PPO2, DQN, and A2C. Fig.~\ref{fig:algocomp} exhibits the trip completion time of these three algorithms as training iterations increase. While all three algorithms end up converging to roughly the same value for trip completion time, we can observe that the A2C algorithm reaches the ideal value faster and exhibits less variance as the training iterations increase. We can conclude from these observations that the A2C algorithm is the best one to use because it produces the most consistent results with the least amount of noise, and is able to reach those results with significantly less training time compared to the other algorithms. We will be using the A2C algorithm to produce our results for the remainder of this paper. 

\subsection {Reward Function Comparison}

Our next experiment is to compare the step and cumulative reward functions to see which one performs better. To do this, we fix the start and end points along with the bandwidth requirement, set the map size to be a 7x7 grid, and then train the agent using each reward function. The agent will be trained while varying the number of high bandwidth cells in the map, the number of maps to be sampled, and their bandwidth allocations will be based on Fig.~\ref{fig:maps}.

\begin{figure}[!h]
    \centering
    \graphicspath{ {./Figures/} }
    \includegraphics [width = 85mm]{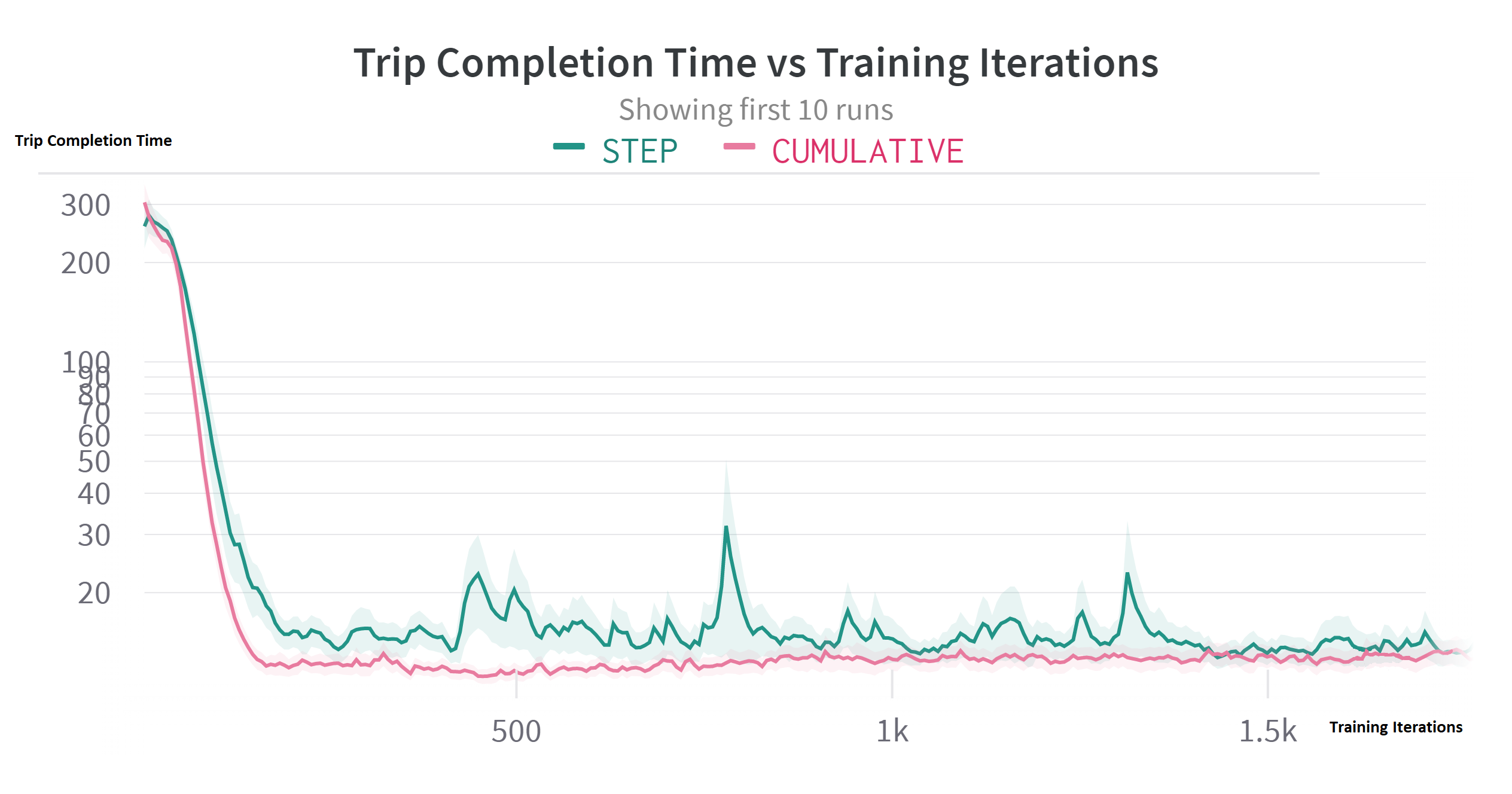}
    \caption{Step Reward vs Cumulative Reward}
    \label{fig:stepvscum}
\end{figure}

Fig.~\ref{fig:stepvscum} shows the trip completion time as a function of training iterations. The plot compares runs across all parameter combinations of the step reward function versus the cumulative reward function. We observe that while the two policies do end up converging to the same values of trip completion time, the cumulative reward function takes less time to converge and seemingly has less spikes along the way. Since this result is consistent across all parameters, we can conclude that the cumulative reward function is more efficient. 

\subsection {Map Parameter Comparisons}

We now take a deeper dive in the data and compare how the agent performs when different map parameters are varied. We can explore the impact of a specific parameter by varying it while fixing all the other parameters. We first compare the agent's performance when the number of maps to be sampled is varied during training. For this experiment, we use the maps in the second row of Fig.~\ref{fig:maps}, setting the number of high bandwidth cells to 4, and the bandwidth requirement to 1 unit of bandwidth.

\begin{figure}[!h]
    \centering
    \graphicspath{ {./Figures/} }
    \includegraphics [width = 85mm]{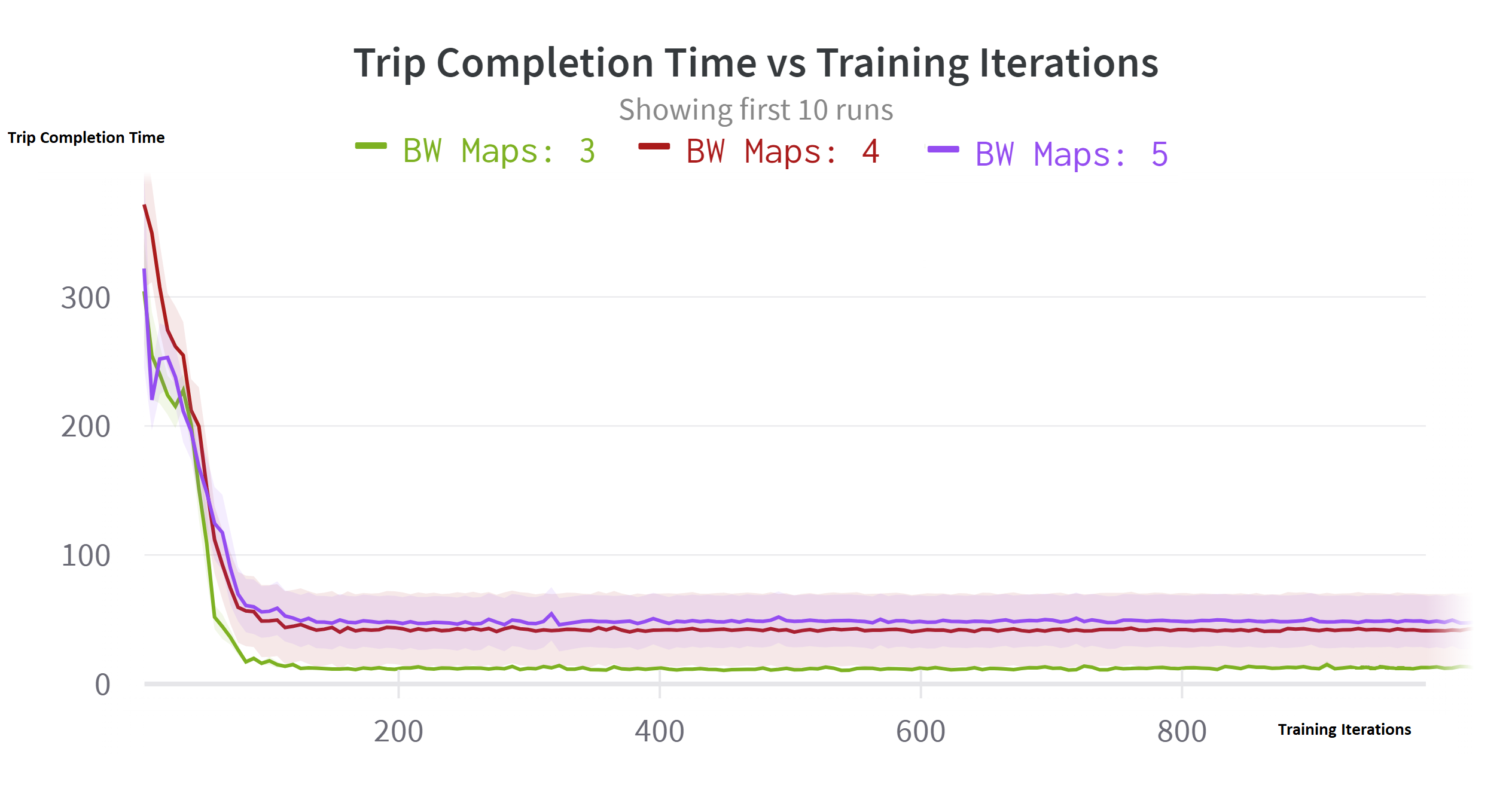}
    \caption{Comparing Different Number of Sampled Maps}
    \label{fig:bwmaps}
\end{figure}

We observe in Fig.~\ref{fig:bwmaps} that there is a slight improvement in the agent’s performance when it samples less maps, each with unique bandwidth allocations, during training. When sampling 3 different maps, the agent converges to a lower value for trip completion time and does so slightly faster than when sampling 4 or 5 different maps.

This observation is in line with expectations: optimization would take longer when there are more variables involved; and the maps selected in the case of selecting 3 maps had low completion times, while the additional maps selected when sampling 4 or 5 maps happened to have longer completion times because of the placement of high-bandwidth cells away from the shortest path. In general, we believe that sampling less maps would require less training; however, we understand that sampling more maps would be required to produce more accurate and applicable results. 


We now compare how the agent performs when the number of high bandwidth cells is varied. For this experiment, the number of maps to be sampled is set to 3, and the bandwidth requirement is set to 1 unit of bandwidth. 

\begin{figure}[!h]
    \centering
    \graphicspath{ {./Figures/} }
    \includegraphics [width = 85mm]{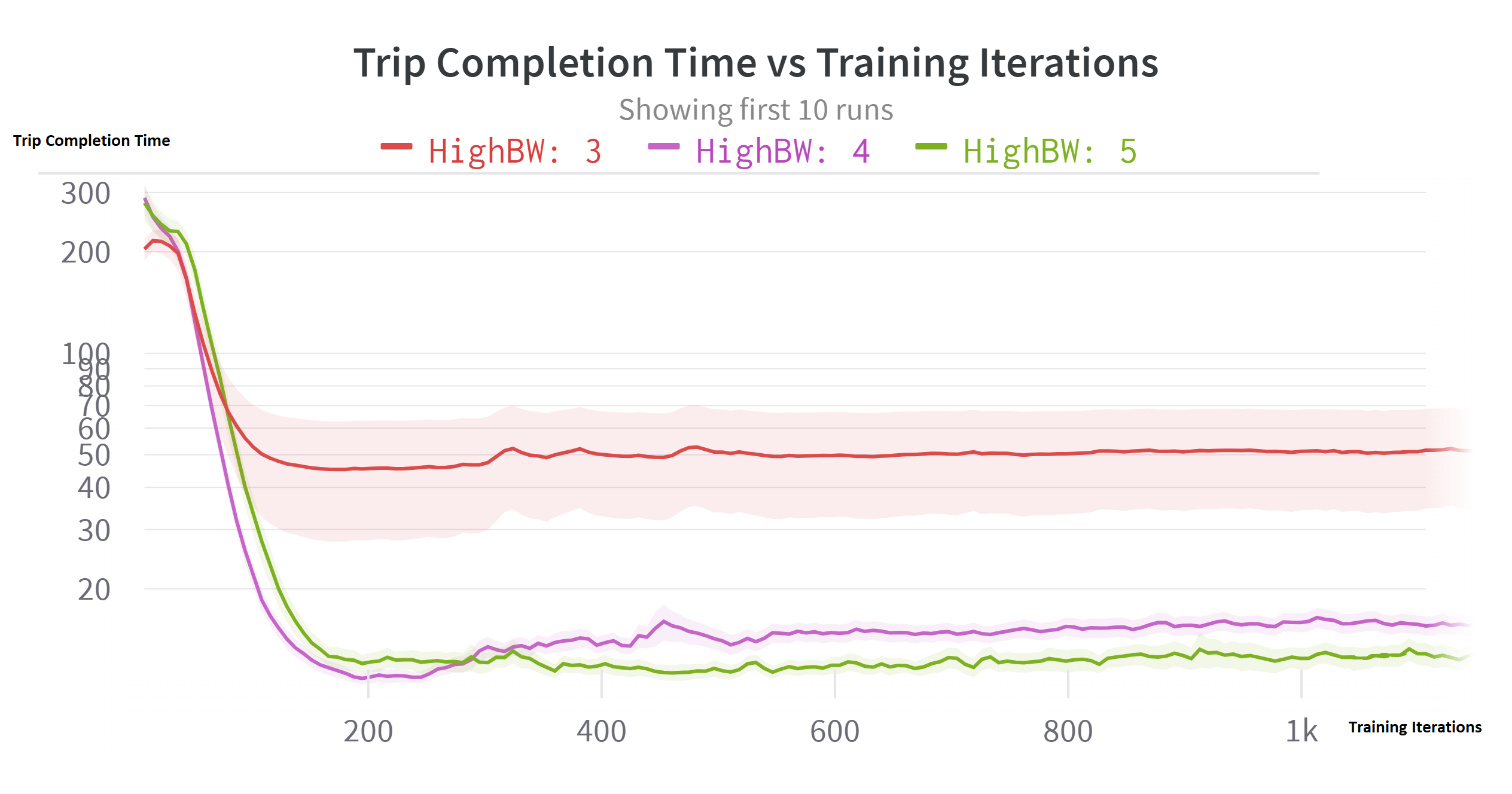}
    \caption{Comparing Different Number of High Bandwidth Cells}
    \label{fig:highbw}
\end{figure}

We observe that the plots in Fig.~\ref{fig:highbw} converge faster when there are more high bandwidth cells in the map. These results are in line with expectations, since maps with more high bandwidth cells open up more possible paths for the agent to take. Traversing through maps with a higher number of high bandwidth cells leads to faster completion times. 

We now compare how the agent performs when the bandwidth requirement is varied. For this experiment, the number of maps to be sampled is set to 3, and the number of high bandwidth cells is set to 5. 

\begin{figure}[!h]
    \centering
    \graphicspath{ {./Figures/} }
    \includegraphics [width = 85mm]{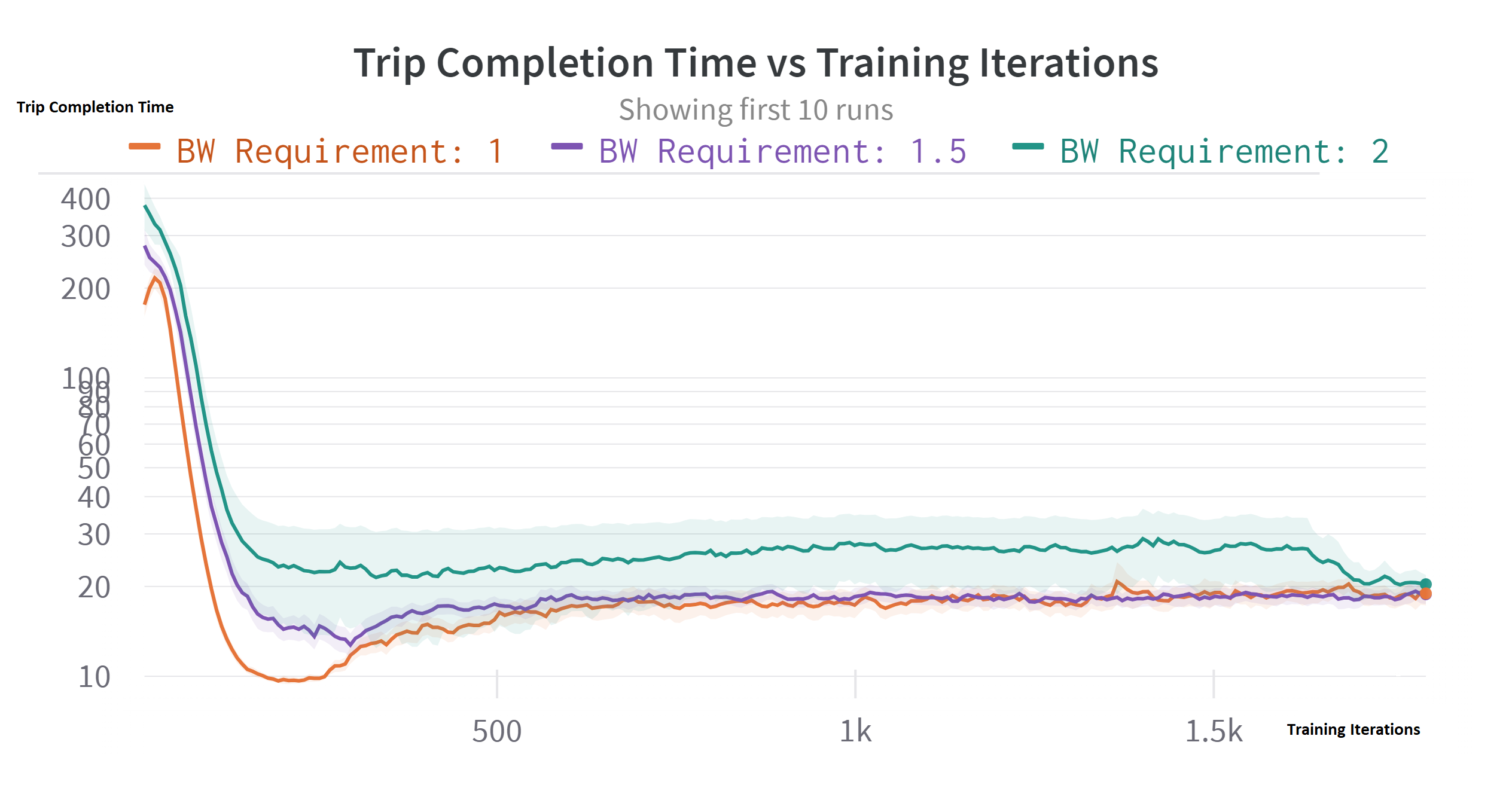}
    \caption{Comparing Different Bandwidth Requirements}
    \label{fig:bwreq}
\end{figure}

We observe in Fig.~\ref{fig:bwreq} that the plots converge faster when the bandwidth requirement is lower. These results are in line with expectations; when an agent has a lower bandwidth requirement, it can achieve the bandwidth goal with less visits to a high bandwidth cell, and can subsequently reach the end destination faster.

\subsection {Baseline Comparisons}

Next, we compare the performance of our trained agent to baselines such as a bandwidth-unaware agent and a traffic-unaware agent. To do this, we sampled three different maps with four high bandwidth cells each during training. 

\begin{figure}[!h]
    \centering
    \graphicspath{ {./Figures/} }
    \includegraphics[width=85mm] {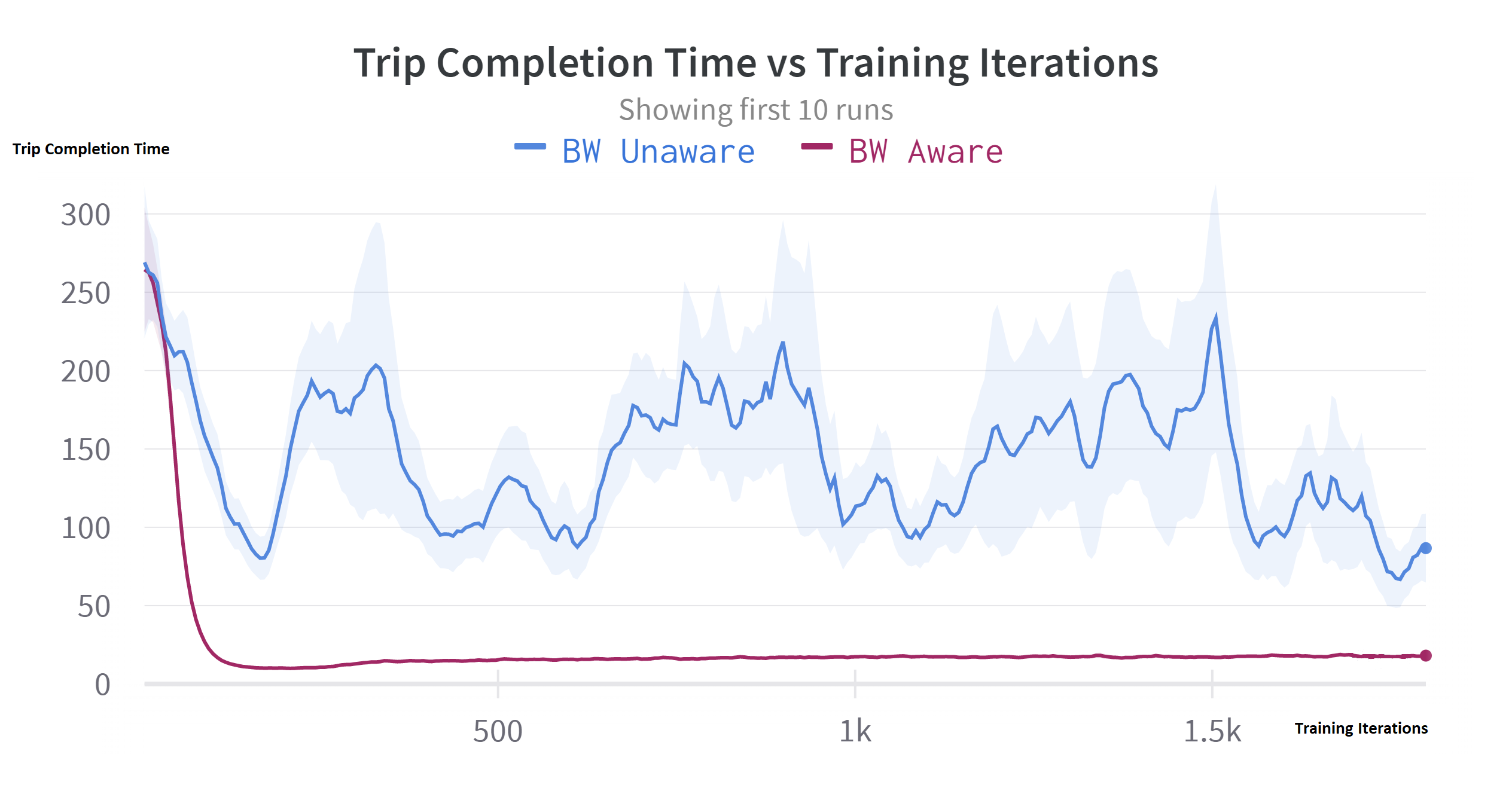}
    \caption{Bandwidth Aware vs Bandwidth Unaware}
    \label{fig:bwaware}
\end{figure}

The bandwidth-unaware agent is an agent that is not being rewarded for visiting a high bandwidth cell, and only rewarded for completing the trip. We can observe from Fig.~\ref{fig:bwaware} that the bandwidth-unaware agent's plot looks considerably different than the bandwidth-aware agent's plot. The bandwidth-unaware agent isn't learning to successfully complete the trip in a consistent manner. These results were in line with our expectations; the bandwidth-unaware agent doesn't learn or register where the high bandwdith locations are, so it would naturally take a longer time for the agent to find those high bandwidth cells and reach the destination.

\begin{figure}[!h]
    \centering
    \graphicspath{ {./Figures/} }
    \includegraphics[width=90mm] {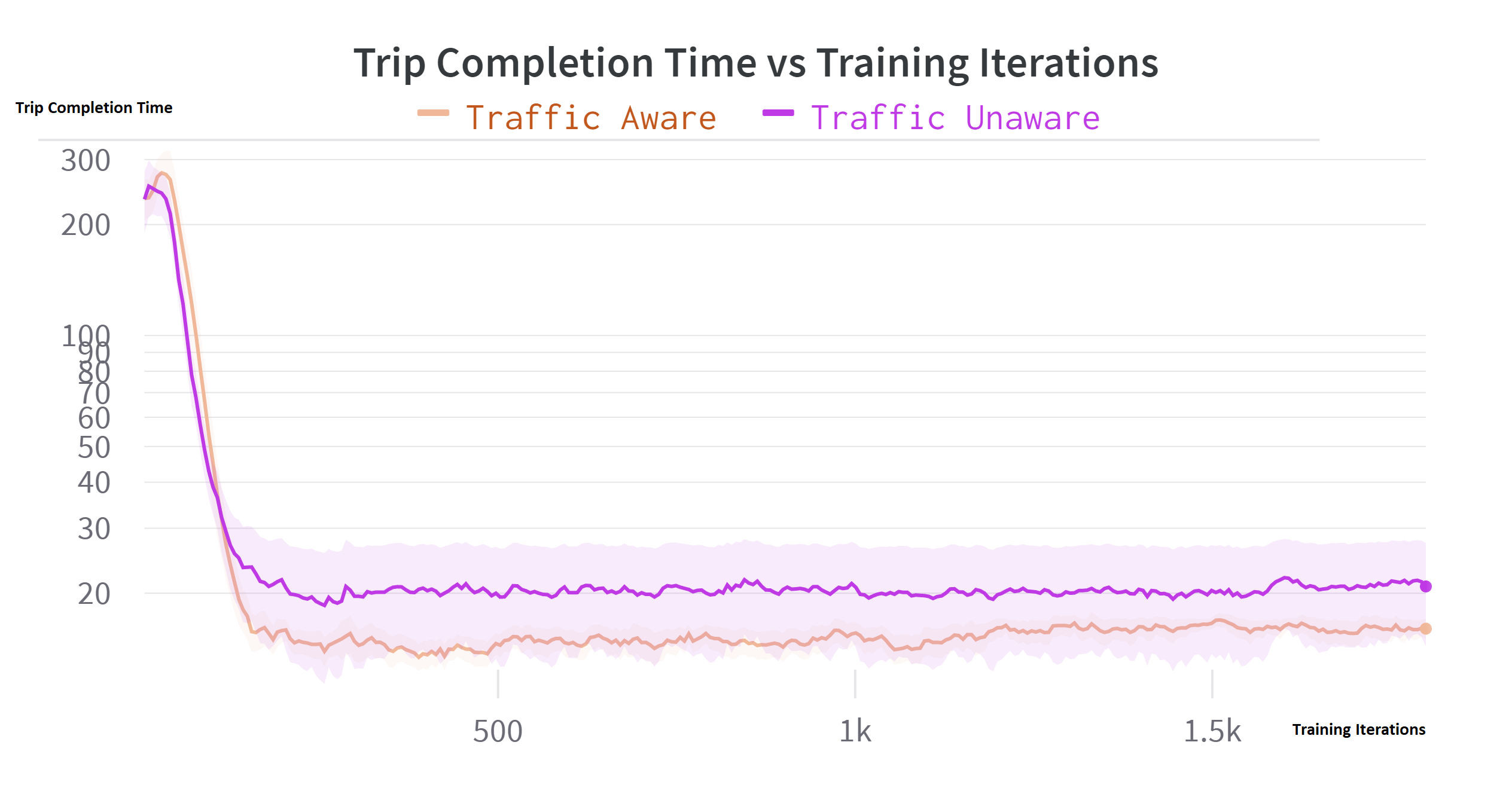}
    \caption{Traffic Aware vs Traffic Unaware}
    \label{fig:trafficaware}
\end{figure}

Fig.~\ref{fig:trafficaware} compares the traffic-unaware agent with the traffic-aware agent. The traffic unaware agent is an agent that is not being given a negative reward, or punishment, that is scaled off the traffic density in each cell. This way, the agent is traversing the map as if traffic is not a factor. We observe that while the traffic-unaware agent does converge slightly earlier than the traffic-aware agent, it also converges at a higher value. We can tell that the traffic-aware agent performs better, since it is optimized for completion time. On the other hand, the traffic-unaware agent is optimized for shortest hop count; it cannot sense the traffic values in the path it is taking, so it converges quicker but completes the episodes in a slower time as a result. 

\section{Conclusion and Future Work} \label{conclusion}

In this paper, we addressed the problem of route planning for autonomous vehicles in urban areas accounting for both drive time and data transfer needs. With the help of real life traffic data, we proposed a route planning solution using RL that aims to prioritize high bandwidth roads to meet the data transfer requirements before reaching the destination, while simultaneously taking the route with the least amount of traffic.

We first compare the performance of three different RL algorithms on the environment and conclude that the Advantage Actor Critic (A2C) algorithm performs better than the Proximal Policy Optimization (PPO2) algorithm and the Deep Q Network (DQN) algorithm. We then compare the step and cumulative reward functions and determine that the cumulative reward function, a function that rewards the agent only after transferring all its data, is more efficient and converges faster than the step reward function. We also compared our RL algorithm with relevant baselines such as bandwidth-unaware and traffic-unaware agents, and our findings were generally in line with expectations. The bandwidth-unaware agent performed significantly worse, since the agent doesn't know or register where the high bandwidth locations are. The traffic-unaware agent does converge slightly faster than our RL algorithm because it doesn't have to worry about finding the path with least traffic, but consequently takes a longer time to complete the trip than our proposed RL algorithm. 

In this paper, we used the traffic data for downtown San Francisco since its streets are already arranged in a grid that closely resembles our OpenAI gym grid environment. It was important to do this to have as realistic of a model as possible. If not, problems would arise from having to deal with highways, irregular stops from traffic lights and stop signs, and roads that are not necessarily vertical or horizontal that wouldn't span across all the cells in our grid. The next step to developing this work would be to create a more generalized environment that would address these issues. We also used binary bandwidth values to model our environment: roads either had bandwidth or didn't. This solution could be generalized for using a continuous range of bandwidth values, should that data be available, and would be a great way to make the model more realistic. 

In our experiments, we fixed the start and end points and used the same set of maps during training. To develop this work, we could average out the set of maps chosen and also explore maps with varying start and end points. This would lead to more variance in results, but would be important to ensure generalizable results. Another consideration for future work would be for latency constraints on data transmission. Having the agent try and find a route that completes data transmission within a certain amount of time would fine tune the motivation behind exploring cellular data transfer in autonomous vehicles.

\section*{Acknowledgment}

Special thanks to Kuwait University for funding the doctoral fellowship of Yousef AlSaqabi and supporting the research reported in this paper. Traffic data was retrieved from Uber Movement, (c) 2022 Uber Technologies, Inc.~\cite{Uber}

\bibliographystyle{unsrt}
\bibliography{Bib.bib}

\end{document}